\documentclass[lettersize,journal]{IEEEtran}
\usepackage{amsmath,amsfonts}
\usepackage{algorithmic}
\usepackage{algorithm}
\usepackage{array}
\usepackage[caption=false,font=normalsize,labelfont=sf,textfont=sf]{subfig}
\usepackage{textcomp}
\usepackage{stfloats}
\usepackage{url}
\usepackage{verbatim}
\usepackage{graphicx}
\usepackage{cite}
\usepackage{booktabs}
\usepackage{multirow}
\usepackage{pifont}
\usepackage{bm}
\usepackage{bbding}
\usepackage{graphicx}
\usepackage{tabularx}
\usepackage{booktabs}
\usepackage{algorithm}
\usepackage{algorithmic}
\usepackage{tikz}
\usetikzlibrary{shapes, arrows}
\usepackage{hyperref}
\begin{document}

\title{Text-Driven Causal Representation Learning for Source-Free Domain Generalization}

\newcommand{\zhou}[1]{\textcolor{red}{#1}}
\author{Lihua Zhou, Mao Ye, ~\IEEEmembership{Senior Member,~IEEE,}
Nianxin Li, Shuaifeng Li, Jinlin Wu, Xiatian Zhu, \\
Lei Deng, Hongbin Liu, Jiebo Luo,~\IEEEmembership{Fellow,~IEEE,}
Zhen Lei$^*$, ~\IEEEmembership{Fellow,~IEEE}
\thanks{
\IEEEcompsocthanksitem Lihua Zhou, Jinlin Wu and Hongbin Liu are with the Centre for Artificial Intelligence and Robotics, Hong Kong Institute of
Science and Innovation, Chinese Academy of Sciences, Hong Kong, China.
Email: lihuazhou120@gmail.com, jinlin.wu@cair-cas.org.hk, liuhongbin@ia.ac.cn
\IEEEcompsocthanksitem Mao Ye, Nianxin Li and Shuaifeng Li are with School of Computer Science and Engineering, University of Electronic Science and Technology of China, Chengdu 611731, China. E-mail: maoye@uestc.edu.cn,  linianxin1220@gmail.com, hotwindlsf@gmail.com
\IEEEcompsocthanksitem Xiatian Zhu is with Surrey Institute for People-Centred Artificial Intelligence, CVSSP, University of Surrey, Guildford, UK. E-mail: xiatian.zhu@surrey.ac.uk
\IEEEcompsocthanksitem {Lei Deng} is with the School of Electronics and Information Engineering,
Shenzhen University, Shenzhen, China. E-mail: ldeng.sjtu@gmail.com
\IEEEcompsocthanksitem {Jiebo Luo} is with the University of Rochester and performed this work while on sabbatical leave at the Hong Kong Institute of Science and Innovation.
\IEEEcompsocthanksitem Zhen Lei is with the School of Artificial Intelligence, University of Chinese Academy of Sciences (UCAS), Beijing 100049, China;
the Centre for Artificial Intelligence and Robotics, Hong Kong Institute of
Science and Innovation, Chinese Academy of Sciences, Hong Kong, China.
Email: zhen.lei@ia.ac.cn
\IEEEcompsocthanksitem * corresponding author.
}
}

\markboth{Journal of \LaTeX\ Class Files,~Vol.~14, No.~8, August~2021}%
{Shell \MakeLowercase{\textit{et al.}}: A Sample Article Using IEEEtran.cls for IEEE Journals}


\maketitle

\begin{abstract}
Deep learning often struggles when training and test data distributions differ. Traditional domain generalization (DG) tackles this by including data from multiple source domains, which is impractical due to expensive data collection and annotation. Recent vision-language models like CLIP enable source-free domain generalization (SFDG) by using text prompts to simulate visual representations, reducing data demands. However, existing SFDG methods struggle with domain-specific confounders, limiting their generalization capabilities. To address this issue, we propose TDCRL (\textbf{T}ext-\textbf{D}riven \textbf{C}ausal \textbf{R}epresentation \textbf{L}earning), the first method to integrate causal inference into the SFDG setting. TDCRL operates in two steps: first, it employs data augmentation to generate style word vectors, combining them with class information to generate text embeddings to simulate visual representations; second, it trains a causal intervention network with a confounder dictionary to extract domain-invariant features. Grounded in causal learning, our approach offers a clear and effective mechanism to achieve robust, domain-invariant features, ensuring robust generalization. Extensive experiments on PACS, VLCS, OfficeHome, and DomainNet show state-of-the-art performance, proving TDCRL’s effectiveness in SFDG.
\end{abstract}


\section{Introduction}
\IEEEPARstart{D}{eep} learning often struggles with domain shifts, where test data distributions deviate from training data, leading to degraded performance on unseen domains \cite{zhou2022domain,wang2022generalizing,wei2024multi,liu2025rotation,xu2025incorporating,zhang2025unified,xu2025unraveling}. 
Traditional Domain generalization (DG) \cite{arjovsky2019invariant,zhoudomain} addresses this by collecting multiple source domains to enhance robustness, 
yet this process typically involves extensive image collection and annotation from diverse domains, which is resource-intensive, time-consuming, and often impractical in real-world applications.

\begin{figure}[t]
  \begin{center}
  \includegraphics[width=\linewidth]{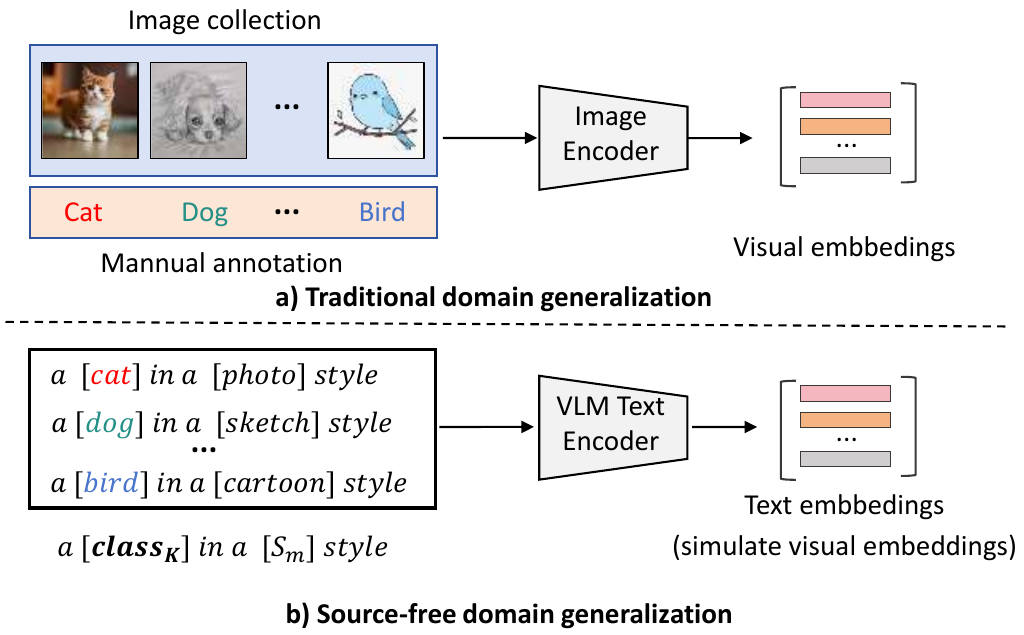}
  \end{center}
     \caption{Comparison of DG and SFDG. (a) DG relies on expensive image collection and manual annotation across multiple source domains, a resource-intensive and time-consuming process. (b) SFDG exploits vision-language models, which align text and visual embeddings in a shared feature space, enabling the simulation of visual embeddings through text prompts. By using predefined text templates to generate diverse text prompts, which are mapped to text embeddings via the vision-language model’s text encoder to simulate visual embeddings, SFDG avoids the need for extensive image datasets, offering a scalable and efficient alternative for robust generalization.}
  \label{fig:0}
\end{figure}

Recent advances in vision-language models (VLMs), such as CLIP, have enabled the alignment of text and image embeddings in a shared space \cite{radford2021learning,jia2021scaling,yang2022vision,lavoiemodeling,zhang2024vision}, making it possible to simulate image using text prompts. This development has paved the way for source-free domain generalization (SFDG), which eliminates the need for collecting and annotating large image datasets \cite{cho2023promptstyler,zhang2024promptta,tang2025dpstyler,xu2025batstyler}. By using simple text-based simulations, SFDG provides a highly efficient and scalable alternative, significantly reducing the time and effort required while maintaining model robustness, as illustrated in Fig. \ref{fig:0}.

\begin{figure}[t]
  \begin{center}
  \includegraphics[width=\linewidth]{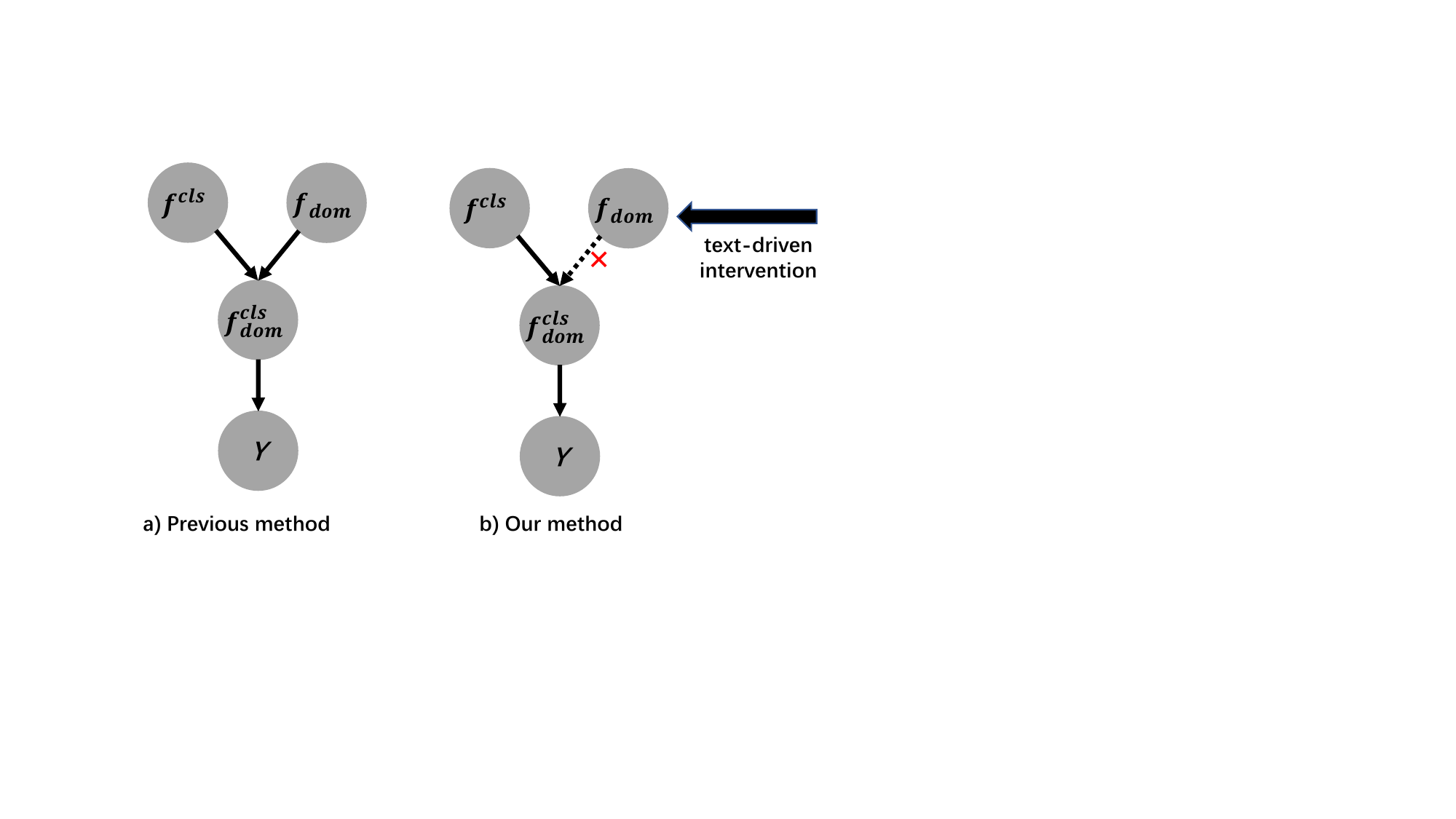}
  \end{center}
     \caption{When training a classifier using observed $\bm{f}^{cls}_{dom}$: (a) Existing SFDG methods often fail to learn the relationship between domain-invariant features $\bm{f}^{cls}$ and labels $Y$ due to the influence of domain-specific confounders $\bm{f}_{dom}$. This results in suboptimal performance when generalizing across different domains. (b) In contrast, our proposed TDCRL method employs causal learning to intervene on the style $\bm{f}_{dom}$, effectively removing its influence. By doing so, we model the direct relationship between domain-invariant features $\bm{f}^{cls}$ and labels $Y$, ensuring robust and generalized performance across various environments. }
  \label{fig:1}
\end{figure}

Currently, SFDG methods can be divided into two categories. The first category employs prompt tuning-based strategies \cite{cho2023promptstyler,zhang2024promptta,xu2025batstyler}, where various style word vectors are learned and combined with class names to generate text embeddings, then a classifier is directly trained based on these text embeddings. Despite its effectiveness, this approach faces challenges when test environments differ from the learned styles, potentially leading to performance degradation. 
The second category applies adapter-based strategies \cite{tang2025dpstyler}, using data augmentation techniques like mixup and random sampling to create varied style word vectors and build text embeddings from them. {This method includes a style removal module that uses entropy to remove domain-specific features, hoping to boost performance across domains. However, relying only on entropy cannot ensure the model learns domain-invariant features, particularly when test settings vary greatly.} In summary, existing methods, especially those based on prompt tuning, are easily affected by domain confounders, limiting their generalization, as shown in Fig.~\ref{fig:1}(a).

To overcome these challenges, causal learning offers a practical solution \cite{pearl2000models,scholkopf2022causality}. Previous DG methods have successfully used causal inference to identify causal relationships, remove domain-specific confounders, and extract domain-invariant features, enhancing model robustness and interpretability. However, in the context of SFDG, this approach has not been explored. 
{In this work, we propose a causal intervention framework to eliminate domain-specific confounders, enabling domain-invariant feature modeling tied to category information and labels with theoretical guarantees for improved generalization, as shown in Fig. \ref{fig:1}(b).
Moreover, unlike DG, where causal intervention often require complex training to simulate style transformations \cite{lv2022causality,ouyang2022causality}, SFDG achieves efficient interventions through simple text prompt modifications. For example, by altering the text prompt from \texttt{"a cat in a \textbf{sketch} style"} to \texttt{"a cat in a \textbf{cartoon} style"}, a causal intervention is achieved that isolates the effect of style variation while keeping class information unchanged. This streamlined intervention simplifies style transformations and enhances model training efficiency.}

To tackle the source-free domain generalization challenge, we propose TDCRL (Text-Driven Causal Representation Learning), a method with two key steps. In the first step, we create varied style word vectors through data augmentation, which are then combined with class names to generate text embeddings using CLIP’s text encoder to simulate image data. In the second step, we use causal learning to obtain domain-invariant features by applying causal interventions to the text embeddings with a confounder dictionary. This step trains a causal intervention network to adjust the text embeddings using the confounder dictionary’s entries, and the resulting domain-invariant features train a classifier. At test time, input images go through CLIP’s visual encoder to generate visual features, which are then adjusted by the causal intervention network using the confounder dictionary, before the trained classifier makes predictions. Our contributions are as follows:  
\begin{enumerate}  
\item We introduce TDCRL, the first method to employ causal inference in the SFDG setting, which uses a confounder dictionary and causal interventions to learn domain-invariant features with clear theoretical support.  
\item We utilize TDCRL, combined with text-based interventions, to simplify the creation of varied styles and obtain domain-invariant features, providing a practical advantage over traditional DG methods that require complex image modifications.  
\item We perform extensive experiments on multiple benchmark datasets to demonstrate that TDCRL achieves state-of-the-art performance, validating its effectiveness and robustness in practical applications.
\end{enumerate}

\section{Related Work}
\label{sec:related}
\noindent\textbf{Domain Generalization.} DG aims to train models using data from multiple source domains, thereby achieving out-of-distribution generalization and enhancing robustness in unseen target domains. Existing DG methods can be categorized into four primary types:
The first type employs domain alignment strategies, focusing on minimizing discrepancies between source domains through statistical measures such as moment matching \cite{muandet2013domain,erfani2016robust}, contrastive loss \cite{motiian2017unified,yoon2019generalizable}, KL divergence \cite{wang2021respecting,li2020domain}, and maximum mean discrepancy \cite{li2018domain}, and adversarial learning techniques \cite{li2018deep,shao2019multi,rahman2020correlation} to achieve domain-invariant representations.
The second type utilizes meta-learning strategies \cite{li2018learning,balaji2018metareg,du2020metanorm}, which involve splitting the training data into meta-training and meta-testing sets. This approach optimizes the model on the meta-training set to improve performance on the meta-testing set, effectively extracting meta-knowledge from multiple source domains to enhance generalization to new domains.
The third type applies data augmentation strategies \cite{shi2020towards,volpi2019addressing,volpi2018generalizing,qiao2020learning}, aiming to diversify training samples through techniques like image transformations and task-adversarial gradients, thus improving the robustness and adaptability of models.
The fourth type leverages self-supervised learning methods \cite{carlucci2019domain,bucci2021self,wang2020learning}, designing pre-training tasks that exploit the intrinsic structure of data without requiring labels. This approach facilitates learning useful representations, thereby enhancing the versatility and flexibility of models.

\noindent\textbf{Vision-Language Models.} 
VLMs integrate visual and textual data into a shared feature space, leveraging large-scale image-text pairs to enable zero-shot classification and cross-modal understanding \cite{radford2021learning,jia2021scaling,yaofilip,chenpali}. Pioneered by CLIP \cite{radford2021learning}, VLMs use contrastive learning to align image and text embeddings, achieving robust generalization across domains without task-specific fine-tuning. Subsequent works like ALIGN \cite{jia2021scaling} scale this approach with over one billion noisy image-text pairs, while FILIP \cite{yaofilip} enhances fine-grained semantic alignment via a late-interaction contrastive mechanism. PaLI \cite{chenpali} jointly scales vision and language components across diverse parameter sizes. These advancements in VLMs enable text embeddings to simulate visual representations, facilitating source-free domain generalization as explored in our work.

\noindent\textbf{{Causal Inference.}} Causal inference aims to uncover causal relationships beyond mere correlations, utilizing tools such as do-operations and structural causal models to distinguish true causes from confounders \cite{pearl2000models,scholkopf2022causality,xue2024integrating,liu2025causal,dong2025intersecting}. This framework provides actionable insights by modeling interventions, setting it apart from traditional predictive methods. In computer vision, causal methods mitigate spurious correlations, such as background biases in image classification \cite{arjovsky2019invariant} and dataset shifts \cite{jones2024causal,wang2020visual}. In reinforcement learning, these methods enhance policy robustness by identifying causal action-outcome links \cite{zhang2020causal,buesing2018woulda}, while in natural language processing, they improve reasoning by addressing confounding variables \cite{feder2022causal}.

In DG, causal inference plays a pivotal role in learning domain-invariant features to address distribution shifts \cite{arjovsky2019invariant,mahajan2021domain,lv2022causality,miao2022domain,lu2021invariant,mao2022causal,yin2024integrating,kong2022partial}.
{IRM \cite{arjovsky2019invariant} proposes learning invariant and causal predictors across multiple environments for improved OOD generalization, addressing limitations of ERM.
MatchDG \cite{mahajan2021domain} uses a two-phase causal matching method to learn invariant representations across domains, improving generalization by leveraging perfect matches and contrastive losses.
iCaRL \cite{lu2021invariant} utilizes a novel framework by learning identifiable causal representations using a non-factorized prior and causal discovery in nonlinear settings.
CT \cite{mao2022causal} adopts a causal transportability framework by leveraging neural representations to estimate invariant causal effects.
CMBRL \cite{yin2024integrating} proposes learning invariant predictors using Causal Markov Blanket representations from the latent space for robust generalization.}
Other approaches use do-operations to intervene on domain-specific confounders \cite{lv2022causality,kong2022partial}, or integrate contrastive learning with causality for robust representations \cite{miao2022domain}.
Unlike these methods that rely on image-based training, our TDCRL leverages the advancements in vision-language models to perform text-based causal interventions. This extends its applicability to source-free settings, offering a novel approach to domain generalization.



\noindent\textbf{Source-Free Domain Generalization.} SFDG aims to train models that generalize to target domains without access to labeled source-domain data, addressing the high costs of data collection and annotation in traditional DG. 
Current methods can be divided into two types:
The first category of methods utilizes prompt tuning techniques to learn multiple style word vectors. For instance, 
PromptStyler \cite{cho2023promptstyler} introduces style diversity loss and content consistency loss to maximize diversity while preserving content information. 
Building on PromptStyler,
PromptTA \cite{zhang2024promptta} further incorporates Style Feature Resampling, which dynamically regenerates sampled features during each training epoch to integrate more diverse domain information.
Similarly, BatStyler \cite{xu2025batstyler} extends PromptStyler by introducing Coarse Semantic Generation technology, which aims to filter out redundant fine-grained semantic constraints to enhance style diversity.
The second category employs adapter-based approaches focused on learning domain-invariant features. DPStyler \cite{tang2025dpstyler} initially generates various style words through data augmentation to obtain text embeddings. It then trains a Style Removal Module to eliminate specific style information from the features, yielding domain-invariant features suitable for classification tasks. While effective, these methods often struggle to ensure invariance across highly diverse test domains, motivating our causal inference-based approach.

\section{Method}
\label{sec:method}


\noindent\textbf{Problem Setting.} In this work, we address the challenge of source-free domain generalization using the pre-trained visual-language model CLIP \cite{radford2021learning}. Given a set of class names $\{\texttt{class}_k\}_{k=1}^K$, we aim to build a model that classifies images from unseen target domains accurately, without needing source-domain data. Our goal is to boost model robustness and generalization through Text-Driven causal representation learning.

\begin{figure*}[t]
  \begin{center}
  \includegraphics[width=0.9\linewidth]{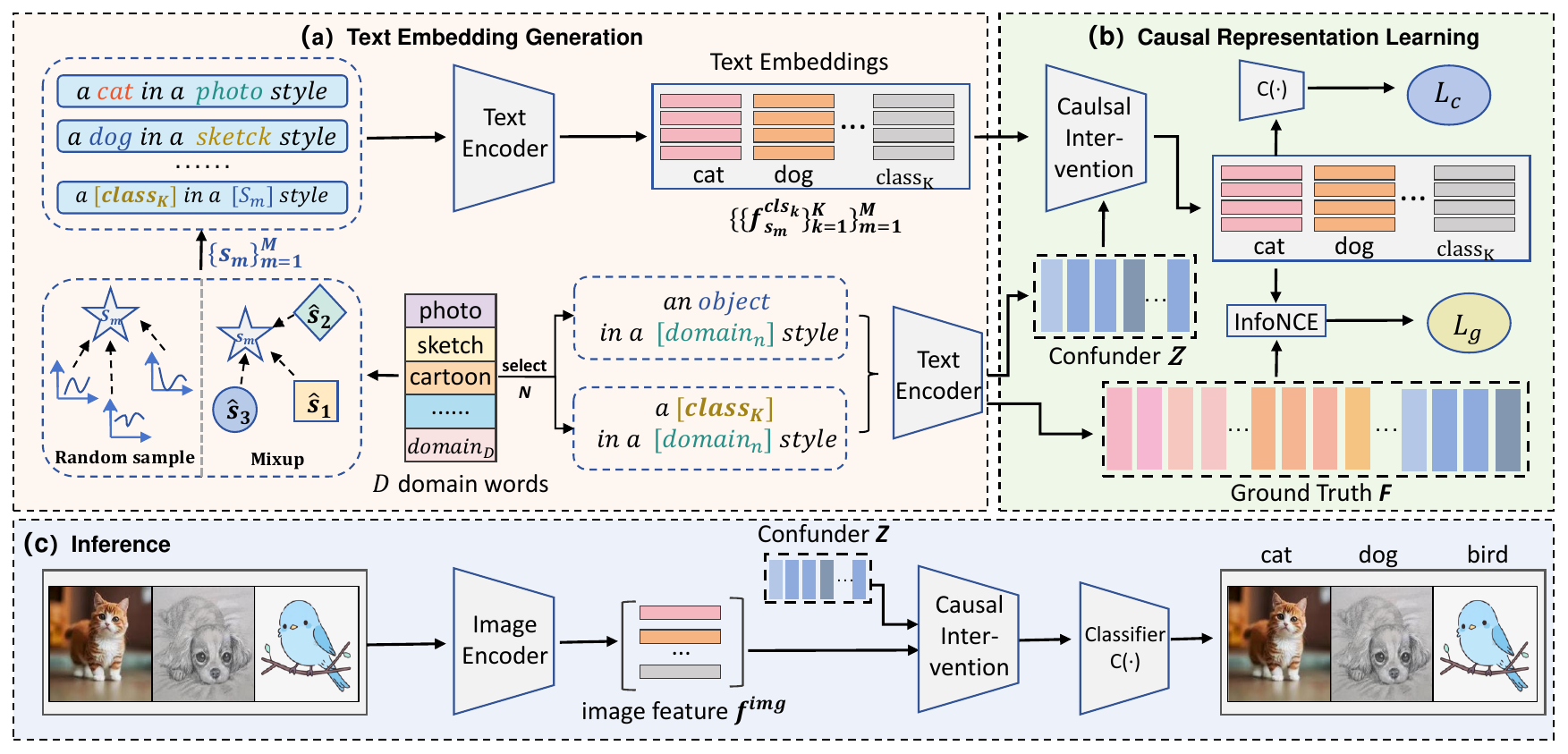}
  \end{center}
     \caption{Overview of our proposed TDCRL. (a) \textit{Text Embedding Generation}: Starting with $D$ predefined domain words, we use mixup and random sampling to generate $M$ style word vectors $\{\bm{s}_m\}_{m=1}^M$ ($M >D$), creating $M \times K$ text embeddings $\{\{\bm{f}_{\bm{s}_m}^{{cls}_k}\}_{k=1}^K\}_{m=1}^M$ via CLIP’s text encoder $T(\cdot)$ as training samples. (b) \textit{Causal Representation Learning}: A confounder dictionary $Z$ of $N$ style embeddings and ground truth $F$, both from $T(\cdot)$, guide a causal intervention network $g(\cdot)$ trained with InfoNCE loss $\mathcal{L}_g$, while a classifier $C(\cdot)$ is trained with cross-entropy loss $\mathcal{L}_C$.  (c) During inference, CLIP’s image encoder $V(\cdot)$ processes target images, followed by $g(\cdot)$ and $C(\cdot)$ for robust classification across unseen domains.}
  \label{fig:overview}
\end{figure*}


\noindent\textbf{Overview.} As shown in Fig.~\ref{fig:overview}, TDCRL leverages CLIP in a two-stage framework to achieve robust generalization without source-domain images. First, we use text embeddings to simulate visual data, expanding a few domain words into diverse styles. Then, we apply causal learning to refine these text embeddings into domain-invariant features and train a classifier. This approach ensures strong performance on unseen target domains, with details in the following sections.


\subsection{Text Embedding Generation}\label{sec:3.1}

Benefiting from VLMs like CLIP \cite{radford2021learning}, which map images and text into a shared feature space, current SFDG methods simulate visual representations using text prompts, eliminating the need to access source domain images.
While recent works like \cite{cho2023promptstyler,zhang2024promptta,xu2025batstyler} use slow and complex prompt tuning to learn multiple style word vectors, our method follows DPStyler \cite{tang2025dpstyler} by adopting a data augmentation strategy to efficiently generate diverse domain styles.

To generate diverse text embeddings, we predefine $D$ domain words $\{\texttt{domain}_d\}_{d=1}^D$ (e.g., {"sketch", "cartoon"}) before training. These words are tokenized into the word vector space to obtain corresponding word vectors $\{\hat{\bm{s}}_d\}_{d=1}^D$. However, since $D$ is typically a small number (e.g., $D=13$ following the setting in \cite{tang2025dpstyler}), directly using these words limits style diversity, failing to capture the full range of domain styles. To address this, we augment the initial $D$ style word vectors to $M$ style word vectors $\{\bm{s}_m\}_{m=1}^M$ ($M > D$, e.g., $M=80$ following the setting in \cite{zhang2024promptta, cho2023promptstyler,xu2025batstyler,tang2025dpstyler}). This augmentation increases the variety of style words, enabling richer representations for downstream feature learning.
We perform this augmentation using a mixup strategy, generating new style embeddings as weighted combinations of the initial vectors:
\begin{equation}
    \bm{s}_m = \sum_{d=1}^{D} w_d \hat{\bm{s}}_d+\epsilon,
\end{equation}
where $w_d$ denotes the corresponding weight coefficient sampled from a beta distribution, satisfying $\sum_{d=1}^{D} w_d = 1$, and $\epsilon$ is a Gaussian noise term that introduces additional variability.


At the beginning of each epoch, we generate $M$ style word vectors $\{\bm{s}_m\}_{m=1}^M$ through $M$ iterations to produce each $\bm{s}_m$. These vectors enable prompts with varied styles and categories, using the template \texttt{a [$\texttt{class}_k$] in a [$\texttt{S}_m$] style}, where $\texttt{class}_k$ is the $k$-th class name and $\texttt{S}_m$ is a placeholder for $\bm{s}_m$. Each prompt is tokenized by CLIP’s tokenizer, mapping tokens to word vectors via embedding lookup, with the word vector of  $\texttt{S}_m$ replaced by $\bm{s}_m$. The resulting text embeddings $\bm{f}_{\bm{s}_m}^{{cls}_k}$ are generated using CLIP’s text encoder $T(\cdot)$. For $K$ categories and $M$ styles, we can obtain $M \times K$ training samples $\{\{\bm{f}_{\bm{s}_m}^{{cls}_k}\}_{k=1}^K\}_{m=1}^M$.

\subsection{Causal Representation Learning}\label{sec:3.2}

\subsubsection{Analysis}\label{sec:3.2.1}



Several existing SFDG methods \cite{zhang2024promptta, cho2023promptstyler,xu2025batstyler} typically minimize the classification error of $P(Y|\bm{f}_{\bm{s}_m}^{{cls}_k})$ to train classifiers using $M \times K$ text embeddings $\bm{f}_{\bm{s}_m}^{{cls}_k}$ generated in the previous section, which combine both category information $\bm{f}^{{cls}_k}$ and domain-specific confounders $\bm{f}_{{\bm{s}_m}}$. However, previous domain generalization studies \cite{arjovsky2019invariant} reveal that $\bm{f}_{{\bm{s}_m}}$ introduces domain noise, causing classifiers to overfit to specific styles rather than learning invariant category traits, thus hindering generalization to unseen domains. 
To achieve this, we aim to model $ P(Y|\bm{f}^{{cls}_k}) $ to capture domain-invariant features. Specifically, we introduce causal inference using the do-operation \cite{pearl2000models} to intervene on $ \bm{f}_{\bm{s}_m}^{{cls}_k} $, modeling $ P(Y|do(\bm{f}_{\bm{s}_m}^{{cls}_k})) $, which mitigates the influence of style confounders and enables the classifier to learn domain-invariant features, as illustrated in Fig. \ref{fig:1}. To this end, we first theoretically establish the equivalence $ P(Y|do(\bm{f}_{\bm{s}_m}^{{cls}_k})) = P(Y|\bm{f}^{{cls}_k}) $.


Specifically, the do-operation \cite{pearl2000models} intervenes on $\bm{f}_{\bm{s}_m}^{{cls}_k}$ by replacing its style component $\bm{f}_{{\bm{s}_m}}$ with predefined styles. To achieve this, we introduce a confounder dictionary $Z=\{\bm{z}_n\}_{n=1}^{N}$, where each $\bm{z}_n$ represents a specific style intervention feature. The intervention process can be formulated as:
\begin{equation}
\begin{aligned}
P(Y|do(\bm{f}_{\bm{s}_m}^{{cls}_k})) &= \sum_{n=1}^{N} P(Y|\bm{f}^{{cls}_k}, \bm{f}_{{\bm{s}_m}} \to \bm{z}_n) P(\bm{z}_n) \\
&= \sum_{n=1}^{N} P(Y|\bm{f}^{{cls}_k}, \bm{z}_n) P(\bm{z}_n).
\label{eq:do}
\end{aligned}
\end{equation}

Next, we apply the Bayesian formula to expand \(P(Y|\bm{f}^{{cls}_k}, \bm{z}_n)\):
\begin{equation}
P(Y|\bm{f}^{{cls}_k}, \bm{z}_n) = \frac{P(Y, \bm{f}^{{cls}_k}, \bm{z}_n)}{P(\bm{f}^{{cls}_k}, \bm{z}_n)},
\label{eq:bayesian}
\end{equation}
where \(P(Y, \bm{f}^{{cls}_k}, \bm{z}_n)\) is the joint probability of \(Y\), \(\bm{f}^{{cls}_k}\), and \(\bm{z}_n\), and \(P(\bm{f}^{{cls}_k}, \bm{z}_n)\) is the joint probability of the conditioned variables.

We assume that \(\bm{f}^{{cls}_k}\) (the category feature) and \(\bm{z}_n\) (the style intervention feature) are independent, as \(\bm{f}^{{cls}_k}\) is domain-invariant and unaffected by style variations, while \(\bm{z}_n\) is a controlled intervention from \(Z\). Thus:
\begin{equation}
P(\bm{f}^{{cls}_k}, \bm{z}_n) = P(\bm{f}^{{cls}_k}) \cdot P(\bm{z}_n),
\label{eq:independent}
\end{equation}
where \(P(\bm{f}^{{cls}_k})\) is the marginal probability of the category feature, and \(P(\bm{z}_n)\) is the probability of selecting style \(\bm{z}_n\).

Substituting Eq.~\eqref{eq:bayesian} and Eq.~\eqref{eq:independent} into Eq.~\eqref{eq:do}, we proceed step-by-step:
\begin{equation}
\begin{aligned}
P(Y|do(\bm{f}_{\bm{s}_m}^{{cls}_k})) &= \sum_{n=1}^{N} \frac{P(Y, \bm{f}^{{cls}_k}, \bm{z}_n)}{P(\bm{f}^{{cls}_k}, \bm{z}_n)} P(\bm{z}_n) \\
&= \sum_{n=1}^{N} \frac{P(Y, \bm{f}^{{cls}_k}, \bm{z}_n)}{P(\bm{f}^{{cls}_k}) \cdot P(\bm{z}_n)} P(\bm{z}_n) \\
&= \sum_{n=1}^{N} \frac{P(Y, \bm{f}^{{cls}_k}, \bm{z}_n)}{P(\bm{f}^{{cls}_k})} \\
&=  \frac{ \sum_{n=1}^{N} P(Y, \bm{f}^{{cls}_k}, \bm{z}_n)}{P(\bm{f}^{{cls}_k})}
\label{eq:4}
\end{aligned}
\end{equation}

Based on the total probability formula, the joint probability \(P(Y, \bm{f}^{{cls}_k}, \bm{z}_n)\) can be summed over all styles in \(Z\):
\begin{equation}
\sum_{n=1}^{N} P(Y, \bm{f}^{{cls}_k}, \bm{z}_n) = P(Y, \bm{f}^{{cls}_k}),
\end{equation}
Substituting this into Eq. \eqref{eq:4}, we have:
\begin{equation}
\begin{aligned}
P(Y|do(\bm{f}_{\bm{s}_m}^{{cls}_k})) 
&= \frac{P(Y, \bm{f}^{{cls}_k})}{P(\bm{f}^{{cls}_k})} = P(Y|\bm{f}^{{cls}_k}).
\end{aligned}
\end{equation}

This proves that $P(Y|do(\bm{f}_{\bm{s}_m}^{{cls}_k})) = P(Y|\bm{f}^{{cls}_k})$.

\subsubsection{Implementation} As shown in Section \ref{sec:3.2.1}, we have proved that by introducing the do-operation in causal learning, we can get $P(Y|do(\bm{f}_{\bm{s}_m}^{{cls}_k})) = P(Y|\bm{f}^{{cls}_k})$. Therefore, in this section, we hope to model this process to train our model. Starting from the initial do-operation, Eq. \eqref{eq:do}, we can obtain the following formula:
\begin{equation}
\begin{aligned}
    P(Y|do(\bm{f}_{\bm{s}_m}^{{cls}_k})) &= \sum_{n=1}^{N} P(Y|\bm{f}^{{cls}_k},  \bm{f}_{{\bm{s}_m}} \xrightarrow{} \bm{z}_n) P(\bm{z}_n)\\
    &=\sum_{n=1}^{N} P(Y|\bm{f}^{{cls}_k}_{\bm{z}_n}) P(\bm{z}_n)\\
    &= \mathbb{E}_{\bm{z}}[P(Y|\bm{f}_{\bm{z}}^{{cls}_k})].
    \label{eq:do1}
\end{aligned}
\end{equation}
Please note that $P(Y|do(\bm{f}_{\bm{s}_m}^{{cls}_k}))$ represents the prediction after intervening on the feature $\bm{f}_{\bm{s}_m}^{{cls}_k}$ using different styles from the confounder dictionary $Z$, $P(Y|\bm{f}_{{\bm{z}_n}}^{{cls}_k})$ denotes the prediction after intervening on $\bm{f}_{\bm{s}_m}^{{cls}_k}$ using a specific style $\bm{z}_n$. Ideally, when $\bm{f}_{\bm{s}_m}^{{cls}_k}$ is intervened with $\bm{z}_n$, the original style information $\bm{f}_{\bm{s}_m}$ in $\bm{f}_{\bm{s}_m}^{{cls}_k}$ will be transformed into $\bm{z}_n$.

In practice, directly observing $\bm{f}^{{cls}_k}$ and $\bm{f}_{\bm{s}_m}$ within $\bm{f}_{\bm{s}_m}^{{cls}_k}$ is infeasible, only the $\bm{f}_{\bm{s}_m}^{{cls}_k}$ can be observed. Consequently, we cannot directly replace $\bm{f}_{\bm{s}_m}$ with $\bm{z}_n$ to simulate the intervention process. To address this, we introduce a neural network $g(\cdot)$, which takes $\bm{f}_{\bm{s}_m}^{{cls}_k}$ and $\bm{z}_n$ as inputs, performing the intervention by transforming $\bm{f}_{\bm{s}_m}$ into $\bm{z}_n$, that is $\bm{f}_{{\bm{z}_n}}^{{cls}_k} = g(\bm{f}_{\bm{s}_m}^{{cls}_k},\bm{z}_n)$.
In addition, we model $P(Y|\bm{f}_{{\bm{z}}}^{{cls}_k})$ using a one-layer linear neural network as a classifier followed by a softmax function:
\begin{equation}
    P(Y|\bm{f}_{{\bm{z}}}^{{cls}_k}) = \text{Softmax}(W \cdot g(\bm{f}_{\bm{s}_m}^{{cls}_k}, \bm{z}) + b),
    \label{eq:linearmodel}
\end{equation}
where $W$ and $b$ are the weights and bias of the classifier, respectively. By substituting Eq. \eqref{eq:linearmodel} back into Eq. \eqref{eq:do1}, we can get the following equation:
\begin{equation}
\begin{aligned}
P(Y|do(\bm{f}_{\bm{s}_m}^{{cls}_k}))= &\mathbb{E}_{\bm{z}}[\text{Softmax}(W \cdot g(\bm{f}_{\bm{s}_m}^{{cls}_k}, \bm{z}) + b)]\\
\overset{\text{NWGM}}{\approx} &\text{Softmax} \left( \mathbb{E}_{\bm{z}} \left[ W \cdot g(\bm{f}_{\bm{s}_m}^{{cls}_k}, \bm{z}) + b\right] \right) \\
=&\text{Softmax} \left(W \cdot  \mathbb{E}_{\bm{z}} \left[ g(\bm{f}_{\bm{s}_m}^{{cls}_k}, \bm{z}) \right] + b\right)
\label{eq:model}
\end{aligned}
\end{equation}
Transitioning from the first to the second line, we apply the Normalized Weighted Geometric Mean approximation \cite{xu2015show}. From Eq.~\eqref{eq:model}, we can see that to carry out this causal inference process, we need to build a confounder dictionary $Z$ and train a causal intervention network $g(\cdot)$ to compute the expected intervened features $\mathbb{E}_{\bm{z}} \left[ g(\bm{f}_{\bm{s}_m}^{{cls}_k}, \bm{z}) \right]$ effectively. {In addition, please note that in the above analysis, the number of items $N$ in the confounder dictionary is usually considered to be infinite, but in the actual implementation process, we can only regard it as a finite number.}

Following the analysis above, we first construct a confounder dictionary by selecting $N$ terms $\{\texttt{domain}_n\}_{n=1}^N$ from the $D$ pre-defined domain words mentioned in the previous section ($N \leq D$, e.g., $N=6$ in our method with an ablation study in \ref{tab:ablation_n}). 
For each term, we create a prompt in the form of \texttt{a object in a $[\texttt{domain}_n]$ style}. These prompts are subsequently fed into the CLIP text encoder to obtain the corresponding style intervention vector $\bm{z}_n$. Through this process, we obtain a confounder dictionary $\bm{Z} = \{\bm{z}_n\}_{n=1}^N$, which encompasses $N$ distinct styles.
Next, the intervention network $g(\bm{f}_{\bm{s}_m}^{{cls}_k}, \bm{z}_n)$ is designed as a neural network comprising multiple fully connected layers. This network takes the text embedding $\bm{f}_{\bm{s}_m}^{{cls}_k}$ and the style intervention vector $\bm{z}_n$ as inputs, outputting a new feature $\bm{f}^{cls}_{\bm{z}_n}$ that incorporates style information intervened from $\bm{f}_{{\bm{s}_m}}$ to $\bm{z}_n$.

\begin{algorithm}[t]
  \caption{Training Procedure}
  \label{alg}
  \textbf{Input}: A pre-trained CLIP model, a set of class names $\{\texttt{class}_k\}_{k=1}^K$ and $D$ predefined domain words $\{\texttt{domain}_d\}_{d=1}^D$, hyperparameters $\tau$, $\lambda$ and $N$.\\
  \textbf{Procedure}: 
  \begin{algorithmic}[1] 
\FOR{each epoch}
    \STATE Generate $M$ style word vectors using data augmentation.
    \STATE Generate $M \times K$ text embeddings based on the generated word vectors.
    \FOR{each iteration}
        \STATE Sample $B$ samples from the $M \times K$ text embeddings.
        \STATE Compute the causal intervention loss $\mathcal{L}_g$.
        \STATE Compute the classification loss $\mathcal{L}_C$.
        \STATE Update the parameters of $g$ and $C$ using $\mathcal{L}_g$ and $\mathcal{L}_C$.
    \ENDFOR
\ENDFOR
  \end{algorithmic}
  \end{algorithm}

{To train the causal intervention network $ g(\cdot) $, we adopt a contrastive learning framework to approximate $ g(\bm{f}_{\bm{s}_m}^{{cls}_k}, \bm{z}_n) $ to $ \bm{f}_{{\bm{z}_n}}^{{cls}_k} $. We supervise this process using a target matrix $ \bm{F} \in \mathbb{R}^{K \times N \times ES} $, where $ ES $ denotes the embedding size. Each element $ \bm{F}[k,n,:] $, representing the ideal embedding for $ \bm{f}_{{\bm{z}_n}}^{{cls}_k} $, is obtained by encoding prompts of the form \texttt{a $[\texttt{class}_k]$ in a $[\texttt{domain}_n]$ style} using CLIP’s text encoder $ T(\cdot) $. For each input $ \bm{f}_{\bm{s}_m}^{{cls}_k} $ and intervention $ \bm{z}_n $, $ g(\cdot) $ is optimized to align its output with the positive sample $ \bm{F}[k,n,:] $, while distancing it from negative samples $ \bm{F}[k,j,:]$ ($ j \neq n $). The InfoNCE loss for each $ \bm{f}_{\bm{s}_m}^{{cls}_k} $ is defined as:}
\begin{equation}
\mathcal{L}_g = -\sum_{n=1}^N \log \frac{\exp(\text{sim}(g(\bm{f}_{\bm{s}_m}^{{cls}_k},\bm{z}_n), \bm{F}[k,n,:]) / \tau)}{\sum_{j=1}^N \exp(\text{sim}(g(\bm{f}_{\bm{s}_m}^{{cls}_k},\bm{z}_n), \bm{F}[k,j,:]) / \tau)}
\end{equation}
where $\text{sim}(\cdot, \cdot)$ is cosine similarity, $\tau$ is the temperature, and the loss aggregates interventions across all $N$ styles in $Z$ for each $\bm{f}_{\bm{s}_m}^{{cls}_k}$. 


We further train a classifier $C(\cdot)$ with the assistance of a causal intervention network $g(\cdot)$, aiming to predict labels from domain-invariant features. As described in Eq.~\eqref{eq:model}, we need to compute the expectation $\mathbb{E}_{\bm{z}} \left[ g(\bm{f}_{\bm{s}_m}^{{cls}_k}, \bm{z}) \right]$. This expectation is approximated by $\frac{1}{N}\sum_{n=1}^N g(\bm{f}_{\bm{s}_m}^{{cls}_k}, \bm{z}_n)$, representing the aggregate effect of all style interventions within the confounder dictionary $\bm{Z}$ used in our method.
The cross-entropy loss for each $\bm{f}_{\bm{s}_m}^{{cls}_k}$ is given by:
\begin{equation}
\mathcal{L}_C = - \log \frac{\exp(C(\frac{1}{N}\sum_{n=1}^N g(\bm{f}_{\bm{s}_m}^{{cls}_k}, \bm{z}_n))[k])}{\sum_{j=1}^K \exp(C(\frac{1}{N}\sum_{n=1}^N g(\bm{f}_{\bm{s}_m}^{{cls}_k}, \bm{z}_n))[j])}
\end{equation}



\noindent\textbf{Train:} Our training process is summarized in {Alg. \ref{alg}}. Before each epoch, we first generate $M$ style word vectors based on data augmentation. Then, we use these word vectors to generate $M \times K$ text embeddings for training. In each iteration, we sample $B$ samples from the $M \times K$ samples to train the causal intervention network $g(\cdot)$ and the classifier $C(\cdot)$ using the losses $\mathcal{L}_C$ and $\mathcal{L}_g$ as follows:
\begin{equation}
\min_{g,C} \mathbb{E}_{\bm{f}_{{\bm{s}_m}}^{{cls}_k}} \mathcal{L}_C + \lambda \mathcal{L}_g,
\end{equation}
where $\lambda$ is a balance hyperparameter.

\noindent\textbf{Inference:} Given an image, we first encode it using the visual encoder of CLIP to obtain the feature representation $\bm{f}^{img}$. We then compute the expected intervened features by applying the causal intervention network $g$ and intervening with the confounder dictionary $\bm{Z}$. We approximate $\mathbb{E}_{\bm{z}} \left[ g(\bm{f}^{img}, \bm{z}) \right]$ as $\frac{1}{N}\sum_{n=1}^N g(\bm{f}^{img}, \bm{z}_n)$. Finally, based on the computed intervened features, we use the classifier $C(\cdot)$ to classify the image and derive the predicted label $\hat{y}$.

\section{Experiments}
\label{sec:exper}


\noindent\textbf{Datasets.}
We evaluate the effectiveness of our proposed method on four widely recognized domain generalization (DG) benchmarks: PACS \cite{li2017deeper}, VLCS \cite{fang2013unbiased}, OfficeHome \cite{venkateswara2017deep}, and DomainNet \cite{peng2019moment}. These datasets are selected for their diverse domains and varying levels of complexity, making them standard choices for assessing robustness to distribution shifts in DG tasks. Specifically, PACS comprises 4 domains (Art Painting, Cartoon, Photo, and Sketch) with 7 classes, challenging models with distinct artistic styles. VLCS includes 4 domains (Caltech, LabelMe, SUN, PASCAL) with 5 classes that test generalization across real-world photographic variations. OfficeHome, with 4 domains (Art, Clipart, Product, and Real World) and 65 classes, providing a broader class diversity relevant to real-world object recognition. Finally, DomainNet, the most comprehensive, spans 6 domains (Clipart, Infograph, Painting, Quickdraw, Real, and Sketch) with 345 classes, posing a significant challenge due to its large scale and domain variability. These benchmarks collectively enable a thorough assessment of our method’s ability to learn domain-invariant features across diverse visual contexts.

\noindent\textbf{Implementation details.} Our experiments are conducted on PyTorch using a single RTX 4070 Ti SUPER GPU, with results averaged over 5 seeds for robustness.
Following \cite{zhang2024promptta,cho2023promptstyler,xu2025batstyler,tang2025dpstyler}, we adopt pre-trained CLIP models \cite{radford2021learning} with ResNet-50 \cite{he2016deep}, ViT-B/16 \cite{dosovitskiyimage} and ViT-L/14 \cite{dosovitskiyimage} as the visual backbone for testing and Transformer \cite{vaswani2017attention} as the text encoder for training in our experiment. The confounder dictionary $\bm{Z} = \{\bm{z}_n\}_{n=1}^N$ uses $N=6$ style world selected from $D=13$ predefined domain words $\{\texttt{domain}_d\}_{d=1}^D$. Before each epoch, we generate $M=80$ style word vectors via data augmentation (mixup and random sampling), producing $M \times K$ text embeddings, where $K$ varies by dataset. The causal intervention network $g(\cdot)$ is a 3-layer linear network. The balance hyperparameter $\lambda$ is set to 3.0 and the temperature hyperparameter $\tau$ is set to 0.1 in all experiments. We set the batch size to 128, train for 60 epochs, and use SGD with an initial learning rate of 0.005, updated via CosineAnnealingLR \cite{loshchilov2016sgdr}. During inference, the test images are pre-processed as in CLIP \cite{radford2021learning} -- resized to 224×224 and normalized.

\noindent\textbf{Domain word details.}
To support diverse visual representations and causal interventions, we predefine a set of \(D = 13\) domain words: "sketch", "cartoon", "photo", "surrealism", "minimalist", "retro", "pixel-art", "collage", "pointillism", "stained-glass", "illustration", "fantasy", and "landscape". These words, used in Section \ref{sec:3.1} for text embedding generation, are chosen to span a variety of artistic and stylistic domains commonly seen in domain generalization tasks, ensuring a rich foundation for style word vectors.

From these \(D = 13\) domain words, we build the confounder dictionary \(Z = \{\bm{z}_n\}_{n=1}^N\) for causal representation learning in Section \ref{sec:3.2}. We select the first \(N = 6\) words in the predefined order: "sketch", "cartoon", "photo", "surrealism", "minimalist", and "retro". This subset balances realistic ("photo"), abstract ("surrealism"), and artistic ("sketch", "cartoon", "minimalist", "retro") styles, providing sufficient diversity for style interventions. Each word is embedded in a prompt like \texttt{a object in a $[\texttt{domain}_n]$ style}, processed by CLIP’s text encoder \(T(\cdot)\) to generate a style intervention vector \(\bm{z}_n\), forming \(Z\) with $N=6$ distinct embeddings.

The choice of \(N=6 \) is guided by two key considerations. First, experiments show that increasing \(N\) beyond 6 yields diminishing performance gains, suggesting that this subset captures most of the beneficial style variety for learning domain-invariant features. Second, as detailed in Section \ref{sec:3.2}, we estimate the expected intervened feature \(\mathbb{E}_{\bm{z}}[g(\bm{f}, \bm{z})]\) as \(\frac{1}{N}\sum_{n=1}^N g(\bm{f}, \bm{z}_n)\); a larger \(N\) increases computational cost, slowing down training and inference without proportional benefits. Thus, \(N = 6\) strikes a practical balance between effectiveness and efficiency. By aligning \(Z\) with a subset of the initial domain words, we ensure consistency between text embedding generation and causal learning, supporting robust generalization as validated in our experiments.



\noindent\textbf{Competitors.} For comprehensive evaluation, we compare TDCRL with CLIP-based state-of-the-art methods, categorized into three groups. The first includes zero-shot CLIP baselines: ZS-CLIP (C) uses $[\texttt{class}_k]$ as the text prompt, and ZS-CLIP (PC) uses \texttt{a photo of a $[\texttt{class}_k]$} for zero-shot prediction \cite{radford2021learning}. The second group leverages source domain data based on CLIP, including CAD \cite{ruanoptimal}, SPG \cite{bai2024soft}, and MIRO \cite{cha2022miro}. The third operates without source data based on CLIP, including PromptStyler \cite{cho2023promptstyler}, PromptTA \cite{zhang2024promptta}, DPStyler \cite{tang2025dpstyler}, and BatStyler \cite{xu2025batstyler}. Performance is evaluated via the top-1 accuracy on unseen domains, reported as the mean accuracy and standard deviation.

\begin{table*}[htbp]
\centering
\caption{Comparison with state-of-the-art (SOTA) methods on four domain generalization benchmarks. Source-free indicates whether source domain data is used during training. $^\dagger$ denotes the results reproduced by DPStyler.}
\label{tab:comparison}
\begin{tabular}{l|c|c|c c c c|c}
\toprule
Method & Venue & Source-free & PACS & VLCS & OfficeHome & DomainNet & Avg. \\
\midrule
\multicolumn{8}{l}{\textit{ResNet-50 with pre-trained weights from CLIP.}} \\ \hline
ZS-CLIP (C) \cite{radford2021learning} & ICML 2021 & \checkmark & 90.6$\pm$0.0 &79.4$\pm$0.0 &67.4$\pm$0.0& 45.9$\pm$0.0 &70.8 \\
ZS-CLIP (PC) \cite{radford2021learning} & ICML 2021 & \checkmark & 90.7$\pm$0.0 &82.0$\pm$0.0& 71.1$\pm$0.0 &46.1$\pm$0.0 &72.5 \\
CAD \cite{ruanoptimal} & ICLR 2022 & \ding{55} & 90.0$\pm$0.6 & 81.2$\pm$0.6 & 70.5$\pm$0.3 & 45.5$\pm$2.1 & 71.8 \\
PromptStyler$^\dagger$ \cite{cho2023promptstyler} & ICCV 2023 & \checkmark & 93.1$\pm$0.4 & 82.2$\pm$0.2 & 71.0$\pm$0.1 & 46.9$\pm$0.2 & 73.3 \\
SPG \cite{bai2024soft} & ECCV 2024&\ding{55} & 92.8$\pm$0.2& 84.0$\pm$1.1 &73.8$\pm$0.5 &\textbf{50.1$\pm$0.2}&75.2 \\
PromptTA \cite{zhang2024promptta} & ARXIV 2024 & \checkmark & {93.8$\pm$0.0} & {83.2$\pm$0.1} & {73.2$\pm$0.1} & {49.2$\pm$0.0} & 74.9 \\
DPStyler \cite{tang2025dpstyler} & TMM 2025 & \checkmark &93.6$\pm$0.2 &83.5$\pm$0.2&72.5$\pm$0.2&48.0$\pm$0.1&74.4\\
BatStyler \cite{xu2025batstyler} & TCSVT 2025 & \checkmark &93.2$\pm$0.2&83.2$\pm$0.2&72.4$\pm$0.4&47.8$\pm$0.1&74.2\\ \hline
\textbf{TDCRL}& \textbf{Ours} & \checkmark & \textbf{94.1$\pm$0.1}&\textbf{84.1$\pm$0.3}&\textbf{74.1$\pm$0.2} &{49.8$\pm$0.2} &\textbf{75.5}\\
\midrule
\multicolumn{8}{l}{\textit{ViT-B/16 with pre-trained weights from CLIP.}} \\ \hline
ZS-CLIP (C) \cite{radford2021learning} & ICML 2021 & \checkmark & 95.6$\pm$0.0 &76.2$\pm$0.0 &79.6$\pm$0.0 &57.4$\pm$0.0 &77.2 \\
ZS-CLIP (PC) \cite{radford2021learning} & ICML 2021 & \checkmark & 96.0$\pm$0.0& 83.0$\pm$0.0& 81.8$\pm$0.0 &57.2$\pm$0.0 &79.5 \\
MIRO \cite{cha2022miro} & ECCV 2022 & \ding{55} & 95.6 & 82.2 & 82.5 & 54.0 & 78.6 \\
PromptStyler$^\dagger$ \cite{cho2023promptstyler} & ICCV 2023 & \checkmark & 96.8$\pm$0.2 & 83.7$\pm$0.5 & 81.8$\pm$0.4 & 56.7$\pm$0.3 & 79.8 \\
SPG \cite{bai2024soft} &ECCV 2024&\ding{55} &97.0$\pm$0.5 &82.4$\pm$0.4 &83.6$\pm$0.4  &60.1$\pm$0.5&80.8\\
PromptTA \cite{zhang2024promptta} & ARXIV 2024 & \checkmark & {97.3$\pm$0.1} & {83.6$\pm$0.3} & {82.9$\pm$0.0} & {59.4$\pm$0.0} & {80.8} \\
DPStyler \cite{tang2025dpstyler} & TMM 2025 & \checkmark& 97.1$\pm$0.1&84.0$\pm$0.4&82.8$\pm$0.1&58.9$\pm$0.1&80.7\\
BatStyler \cite{xu2025batstyler} & TCSVT 2025 & \checkmark&97.3$\pm$0.1&82.7$\pm$0.4&83.7$\pm$0.2&58.5$\pm$0.4 &80.6\\ \hline
\textbf{TDCRL}& \textbf{Ours} & \checkmark &\textbf{97.7$\pm$0.2}&\textbf{84.6$\pm$0.1}&\textbf{83.9$\pm$0.2}&\textbf{60.4$\pm$0.3}&\textbf{81.7}\\
\midrule
\multicolumn{8}{l}{\textit{ViT-L/14 with pre-trained weights from CLIP.}} \\ \hline
ZS-CLIP (C) \cite{radford2021learning} & ICML 2021 & \checkmark & 97.6$\pm$0.0 &77.5$\pm$0.0 &85.7$\pm$0.0 &63.1$\pm$0.0 &81.0 \\
ZS-CLIP (PC) \cite{radford2021learning} & ICML 2021 & \checkmark & 98.3$\pm$0.0 &81.9$\pm$0.0& 86.6$\pm$0.0 &63.0$\pm$0.0 &82.5 \\
PromptStyler$^\dagger$ \cite{cho2023promptstyler} & ICCV 2023 & \checkmark & 98.4$\pm$0.1 & 81.3$\pm$0.3 & 86.4$\pm$0.2 & 62.9$\pm$0.2 & 82.2 \\
PromptTA \cite{zhang2024promptta} & ARXIV 2024 & \checkmark & {98.6$\pm$0.0} & {83.3$\pm$0.3} & {88.5$\pm$0.0} & {65.2$\pm$0.0} & {83.9} \\
DPStyler \cite{tang2025dpstyler} & TMM 2025 & \checkmark &98.4$\pm$0.1&83.2$\pm$0.1&88.0$\pm$0.3&64.7$\pm$0.1&83.6\\
BatStyler \cite{xu2025batstyler} & TCSVT 2025 & \checkmark&98.4$\pm$0.1&82.2$\pm$0.1&87.4$\pm$0.4&64.4$\pm$0.3&83.1 \\ \hline
\textbf{TDCRL}& \textbf{Ours} & \checkmark & \textbf{99.0$\pm$0.1} & \textbf{84.6$\pm$0.2}&\textbf{89.0$\pm$0.2}&\textbf{65.6$\pm$0.1}&\textbf{84.6} \\
\bottomrule
\end{tabular}
\end{table*}

\subsection{Comparisons with State-of-the-art}

\begin{table*}[htbp]
\centering
\caption{Detailed performance of TDCRL across subdomains of PACS, VLCS, Office-Home, and DomainNet. Results are reported as accuracy (\%) with standard deviation.}
\label{tab:detailed}
\begin{tabular}{l|cccc|c}
\toprule
\multicolumn{6}{c}{PACS} \\
\midrule
Backbone & Art Painting & Cartoon & Photo & Sketch & Avg. \\
\midrule
ResNet-50 & 94.6±0.0 & 96.1±0.1 & 100.0±0.0 & 85.8±0.1 & 94.1±0.1 \\
ViT-B/16 & 98.5±0.1 & 99.3±0.1 & 100.0±0.0 & 93.1±0.2 & 97.7±0.2 \\
ViT-L/14 & 99.5±0.1 & 100.0±0.0 & 100.0±0.0 & 96.4±0.0 & 99.0±0.1 \\
\midrule
\multicolumn{6}{c}{VLCS} \\
\midrule
Backbone & Caltech & LabelMe & SUN & PASCAL & Avg. \\
\midrule
ResNet-50 & 100.0±0.0 & 74.2±0.3 & 73.0±0.2 & 89.1±0.2 & 84.1±0.3 \\
ViT-B/16 & 100.0±0.0 & 72.1±0.2 & 75.8±0.3 & 90.5±0.1 & 84.6±0.1 \\
ViT-L/14 & 100.0±0.0 & 71.2±0.1 & 78.0±0.3 & 89.0±0.2 & 84.6±0.2 \\
\midrule
\multicolumn{6}{c}{Office-Home} \\
\midrule
Backbone & Art & Clipart & Product & Real World & Avg. \\
\midrule
ResNet-50 & 74.8±0.2 & 53.7±0.4 & 83.1±0.1 & 84.6±0.1 & 74.1±0.2 \\
ViT-B/16 & 85.1±0.2 & 69.1±0.4 & 90.5±0.1 & 90.7±0.1 & 83.9±0.2 \\
ViT-L/14 & 89.5±0.3 & 77.5±0.2 & 95.1±0.2 & 93.9±0.2 & 89.0±0.2 \\
\end{tabular}

\vspace{0.2cm} 

\begin{tabular}{l|cccccc|c}
\toprule
\multicolumn{8}{c}{DomainNet} \\
\midrule
Backbone & Clipart & Infograph & Painting & Quickdraw & Real & Sketch & Avg. \\
\midrule
ResNet-50 & 57.4±0.1 & 42.4±0.2 & 57.1±0.3 & 11.0±0.2 & 78.2±0.1 & 52.7±0.3 & 49.8±0.2 \\
ViT-B/16 & 72.8±0.4 & 52.0±0.2 & 68.2±0.2 & 18.8±0.5 & 84.4±0.1 & 66.3±0.2 & 60.4±0.3 \\
ViT-L/14 & 79.6±0.1 & 55.3±0.2 & 74.3±0.2 & 24.9±0.1 & 86.4±0.1 & 73.1±0.1 & 65.6±0.1 \\
\bottomrule
\end{tabular}
\end{table*}

Table~\ref{tab:comparison} reports the Top-1 accuracy of TDCRL against CLIP-based state-of-the-art methods on four domain generalization benchmarks (PACS, VLCS, OfficeHome, DomainNet) using ResNet-50, ViT-B/16, and ViT-L/14 backbones. TDCRL consistently achieves the highest average accuracy across all backbones: 75.5\%, 81.7\%, and 84.6\%, outperforming both Source-free methods (e.g., PromptTA: 74.9\%, 80.8\%, 83.9\%) and source-dependent methods (e.g., SPG: 75.2\% on ResNet-50, 80.8\% on ViT-B/16).

For ResNet-50, TDCRL outperforms all methods on PACS (94.1\%, +0.3\% over PromptTA’s 93.8\%), VLCS (84.1\%, +0.1\% over SPG’s 84.0\%), and OfficeHome (74.1\%, +0.3\% over SPG’s 73.8\%). However, on DomainNet, it scores 49.8\%, trailing SPG (50.1\%) by 0.3\%. This minor gap may stem from DomainNet's large class count (345), where multi-source DG methods like SPG leverage richer source data. With ViT-B/16, TDCRL leads on PACS (97.7\%, +0.4\% over PromptTA/BatStyler’s 97.3\%), VLCS (84.6\%, +0.6\% over DPStyler’s 84.0\%), OfficeHome (83.9\%, +0.2\% over BatStyler’s 83.7\%), and DomainNet (60.4\%, +0.3\% over SPG’s 60.1\%). For ViT-L/14, TDCRL excels on PACS (99.0\%, +0.4\% over ZS-CLIP (PC)/PromptTA’s 98.6\%), VLCS (84.6\%, +1.3\% over PromptTA’s 83.3\%), OfficeHome (89.0\%, +0.5\% over PromptTA’s 88.5\%), and DomainNet (65.6\%, +0.4\% over PromptTA’s 65.2\%).

From the Table~\ref{tab:comparison}, we also have the following observations. First, compared to zero-shot CLIP baselines, DG’s fine-tuning yields substantial gains, underscoring DG’s necessity over static prompting.
Second, the performance gap between SFDG methods (e.g., TDCRL, PromptTA) and source-dependent DG methods (e.g., SPG, CAD) narrows significantly across backbones, with TDCRL even surpassing SPG in most cases (e.g., 81.7\% vs. 80.8\% on ViT-B/16). This minimal difference suggests that VLMs like CLIP possess robust generalization capabilities, enabling SFDG to approach or exceed multi-source DG performance without source data. 
Third, TDCRL’s low standard deviations (e.g.,$\pm$0.1 to $\pm$0.3) further highlight its stability compared to higher variances in SPG ($\pm$1.1 on VLCS, ResNet-50), underscoring the effectiveness of causal interventions in mitigating domain-specific noise. The slight DomainNet lag on ResNet-50 (49.8\% vs. 50.1\%) may reflect its vast class diversity (345 classes), where source data aids traditional DG, yet TDCRL’s gains with larger VLMs (e.g., 60.4\% on ViT-B/16) indicate that advanced backbones amplify the ability to extract domain-invariant features, leveraging VLM’s pre-trained knowledge for generalization.

Table~\ref{tab:detailed} supplements the average accuracy results of TDCRL reported in Table \ref{tab:comparison} by providing a detailed breakdown of performance across individual subdomains of PACS, VLCS, Office-Home, and DomainNet. The table lists accuracy (\%) and standard deviation for TDCRL under three backbones—ResNet-50, ViT-B/16, and ViT-L/14—all initialized with pre-trained CLIP weights \cite{radford2021learning}. This subdomain-level detail complements the aggregated results in the Table \ref{tab:comparison}, offering a comprehensive view of TDCRL’s performance across diverse visual styles.

\subsection{Ablation Studies}







\begin{table}[h]
\centering
\caption{Component analysis of causal intervention (CI) on four domain generalization benchmarks using ResNet-50, ViT-B/16, and ViT-L/14 backbones (top to bottom). "w/o CI" denotes directly using cross-entropy loss $\mathcal{L}_C$ for classification.}
\label{tab:component}
\begin{tabular}{l|c c c c}
\toprule
Method & PACS & VLCS & OfficeHome & DomainNet \\ 
\midrule
\multicolumn{5}{c}{\textit{ResNet-50 with pre-trained weights from CLIP.}} \\ \hline
{w/o CI} & 92.3 & 82.0 & 70.7 & 46.5 \\
{TDCRL} & 94.1 & 84.1 & 74.1 & 49.8 \\
\midrule
\multicolumn{5}{c}{\textit{ViT-B/16 with pre-trained weights from CLIP.}} \\\hline
{w/o CI} & 96.3 & 83.5 & 82.1 & 57.8 \\
{TDCRL} & 97.7 & 84.6 & 83.9 & 60.4 \\
\midrule
\multicolumn{5}{c}{\textit{ViT-L/14 with pre-trained weights from CLIP.}} \\\hline
{w/o CI} & 98.3 & 83.0 & 87.3 & 64.4 \\
{TDCRL} & 99.0 & 84.6 & 89.0 & 65.6 \\
\bottomrule
\end{tabular}
\end{table}

\noindent{\bf Component analysis.}  
To assess the contribution of causal intervention (CI) in TDCRL, we ablate the InfoNCE loss $\mathcal{L}_g$, training the causal intervention network $g(\cdot)$ and classifier $C(\cdot)$ directly on text embeddings $\{\bm{f}_{\bm{s}_m}^{{cls}_k}\}$ with cross-entropy loss $\mathcal{L}_C$. Results across PACS, VLCS, OfficeHome, and DomainNet using ResNet-50, ViT-B/16, and ViT-L/14 backbones are shown in Table~\ref{tab:component}. 
Across all backbones, removing CI reduces accuracy by 0.7\%--3.4\%, with larger gains on ResNet-50 (avg. +2.6\%) than ViT-L/14 (avg. +1.5\%), suggesting CI is crucial for mitigating style confounders ($\bm{f}_{dom}$) in smaller models, where raw embeddings are less robust. The improvement validates the necessity of causal intervention for domain-invariant feature extraction.

\begin{table}[h]
\centering
\caption{Cross-modal transferability analysis of style classification and MMD distance on four domain generalization benchmarks using ResNet-50, ViT-B/16, and ViT-L/14 backbones. ``w/o CI'' refers to raw image embeddings \(\bm{f}^{img}\), while ``TDCRL'' denotes operation on intervened features \(\mathbb{E}_z[g(\bm{f}^{img}, z)]\). MMD distances are averaged across domain pairs.}
\label{tab:style_analysis}
\begin{tabular}{l|c c c c}
\toprule
Method & PACS & VLCS & OfficeHome & DomainNet \\ 
\midrule
\multicolumn{5}{c}{\textit{Style Classification Cross-Entropy Loss (\(\mathcal{L}_{CE}\)) with ResNet-50}} \\ \hline
{w/o CI}& 0.42 & 0.41 & 0.50 & 0.53  \\
{TDCRL} & 1.31 & 1.34 & 1.36 & 1.38 \\
\midrule
\multicolumn{5}{c}{\textit{Style Classification Cross-Entropy Loss (\(\mathcal{L}_{CE}\)) with ViT-B/16}} \\ \hline
{w/o CI} & 0.34 & 0.35 & 0.44 & 0.47 \\
{TDCRL} & 1.15 & 1.12 & 1.22 & 1.29 \\
\midrule
\multicolumn{5}{c}{\textit{Style Classification Cross-Entropy Loss (\(\mathcal{L}_{CE}\)) with ViT-L/14}} \\ \hline
{w/o CI}  & 0.26 & 0.23 & 0.25 & 0.29\\
{TDCRL}  & 1.08 & 1.04 & 1.12 & 1.07 \\
\midrule
\multicolumn{5}{c}{\textit{Average MMD Distance with ResNet-50}} \\ \hline
{w/o CI}& 0.65 & 0.61 & 0.72 & 0.75  \\
{TDCRL} & 0.12 & 0.11 & 0.13 & 0.17 \\
\midrule
\multicolumn{5}{c}{\textit{Average MMD Distance with ViT-B/16}} \\ \hline
{w/o CI} & 0.41 & 0.38 & 0.49 & 0.51 \\
{TDCRL} & 0.073 & 0.077 & 0.065 & 0.082 \\
\midrule
\multicolumn{5}{c}{\textit{Average MMD Distance with ViT-L/14}} \\ \hline
{w/o CI} &0.32&0.34&0.38 &0.38\\
{TDCRL}  &0.065&0.070&0.068&0.073\\
\bottomrule
\end{tabular}
\end{table}

\noindent{\bf Cross-modal transferability.}
To validate the cross-modal transferability of the Causal Intervention Network (CIN) in TDCRL, designed to remove domain-specific style information, we analyzed the style classification cross-entropy loss and MMD distance of visual features before and after causal intervention on four domain generalization benchmarks (PACS, VLCS, OfficeHome, DomainNet) using ResNet-50, ViT-B/16, and ViT-L/14 backbones. Training uses text embeddings, while inference applies image embeddings. Table~\ref{tab:style_analysis} shows that style classification loss increases significantly for ViT-B/16 (e.g., from 0.34 to 1.15 on PACS, 0.35 to 1.12 on VLCS, 0.44 to 1.22 on OfficeHome, and 0.57 to 1.29 on DomainNet) post-intervention, indicating that CIN effectively removes style information from visual features. MMD distances between domain pairs decrease substantially for ViT-B/16 (e.g., from 0.41 to 0.073 on PACS, 0.38 to 0.077 on VLCS, 0.39 to 0.065 on OfficeHome, and 0.46 to 0.082 on DomainNet), confirming enhanced domain invariance. These results demonstrate that CIN’s ability to eliminate style information, learned from text embeddings, successfully transfers to image embeddings, supporting robust cross-modal transferability across diverse datasets.

\begin{table}[h]
\centering
\caption{Ablation study on $\mathcal{L}_g$ loss functions with ViT-B/16 backbone across four benchmarks. "L2" denotes L2 loss, "Con" denotes contrastive loss. Results report Top-1 accuracy (\%).}
\label{tab:loss_g}
\begin{tabular}{l|c c c c}
\toprule
Loss & PACS & VLCS & OfficeHome & DomainNet \\
\midrule
L2 & 97.1 & 84.7 & 83.4 & 59.7 \\
Con & 97.7 & 84.6 & 83.9 & 60.4 \\
\bottomrule
\end{tabular}
\end{table}

\noindent{\bf Ablation study on $\mathcal{L}_g$.}
To evaluate the robustness of training the causal intervention network $g(\cdot)$ with different loss functions, we compare the contrastive loss $\mathcal{L}_g$ with an L2 loss alternative across four datasets using the ViT-B/16 backbone. The contrastive loss aligns intervened features with positive samples discriminatively, while the L2 loss, $\mathcal{L}_{L2} = \frac{1}{N} \sum_{n=1}^N \| g(\bm{f}_{\bm{s}_m}^k, \bm{z}_n) - \bm{F}[k, n, :]\|^2_2$, minimizes Euclidean distance. 
Results are shown in Table~\ref{tab:loss_g}. With contrastive loss, TDCRL achieves 97.7\% on PACS (vs. 97.1\% w/ L2, +0.6\%), 84.6\% on VLCS (-0.1\% below 84.7\%), 83.9\% on OfficeHome (+0.5\% over 83.4\%), and 60.4\% on DomainNet (+0.7\% over 59.7\%). The contrastive loss outperforms L2 on most datasets, notably improving generalization on OfficeHome and DomainNet by enforcing discriminative feature separation, crucial for handling diverse styles. On VLCS, L2’s slight edge (84.7\% vs. 84.6\%) may reflect its simpler optimization fitting this dataset’s smaller class set (5 classes). Overall, the differences range from -0.1\% to +0.7\%, with an average absolute change of 0.5\%, indicating that $g(\cdot)$’s training is robust to loss function choice. This stability suggests that the causal intervention mechanism, central to TDCRL, effectively extracts domain-invariant features regardless of the specific loss, highlighting its resilience across diverse domains and reinforcing the reliability of our approach.

\begin{table}[h]
\centering
\caption{Ablation study on $\mathcal{L}_C$ with ViT-B/16 across four benchmarks. "CE" denotes cross-entropy, "Arc" denotes ArcLoss. Results report Top-1 accuracy (\%).}
\label{tab:loss_C}
\begin{tabular}{l|c c c c}
\toprule
Loss & PACS & VLCS & OfficeHome & DomainNet \\
\midrule
Arc & 97.2 & 84.5 & 83.3 & 60.2 \\
CE  & 97.7 & 84.6 & 83.9 & 60.4 \\
\bottomrule
\end{tabular}
\end{table}

\noindent{\bf Ablation study on $\mathcal{L}_C$.}
To evaluate the robustness with different classification loss functions $\mathcal{L}_C$, we compare the cross-entropy loss with ArcFace loss (ArcLoss) \cite{deng2019arcface}, which adds an angular margin to enhance class separation, across four datasets using the ViT-B/16 backbone. 
Results in Table~\ref{tab:loss_C} show minimal variation. CE achieves 97.7\% on PACS (vs. 97.1\% w/ ArcLoss, +0.6\%), 84.6\% on VLCS (+0.1\% over 84.5\%), 83.9\% on OfficeHome (+0.6\% over 83.3\%), and 60.4\% on DomainNet (+0.2\% over 60.2\%). The differences range from 0.1\% to +0.6\%, with an average absolute change of 0.4\%, indicating that $\mathcal{L}_C$’s choice has negligible impact on performance. This small variation underscores TDCRL’s robustness, as the causal intervention framework ensures stable domain-invariant features, rendering $C(\cdot)$ training resilient to loss function variations.

\begin{figure}[t]
\centering
\includegraphics[width=1\linewidth]{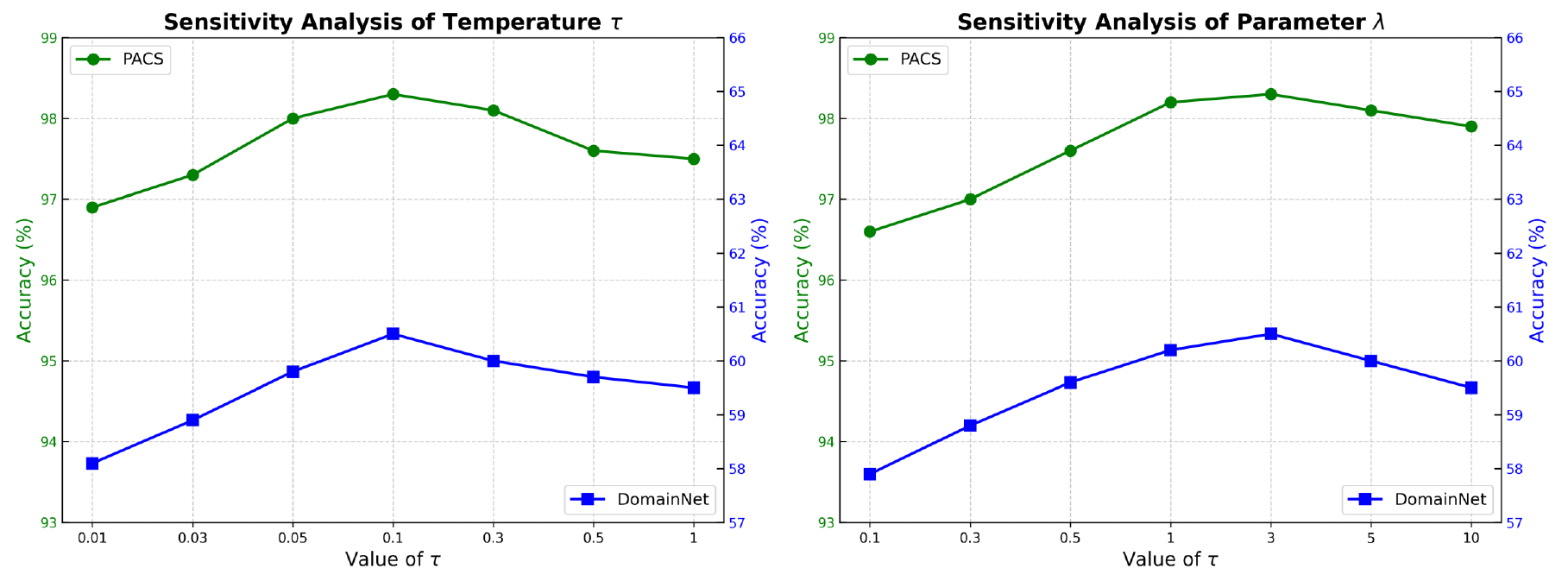}
\caption{Sensitivity analysis with respect to $\tau/\lambda$ on PACS and DomainNet using ViT-B/16 as the visual backbone.}
\label{fig:3}
\end{figure}

\noindent{\bf Sensitivity analysis of $\tau$ and $\lambda$.} 
To investigate the sensitivity of TDCRL to hyperparameters $\tau$ (temperature in $\mathcal{L}_g$) and $\lambda$ (balancing $\mathcal{L}_g$ and $\mathcal{L}_C$), we conduct experiments on PACS and DomainNet using the ViT-B/16 backbone. 
Results, illustrated in Fig.~\ref{fig:3}, reveal that accuracy initially rises and then declines as these hyperparameter values increase. 
Moreover, TDCRL demonstrates robustness to variations in $\tau$ and $\lambda$, maintaining strong performance across diverse settings. 
This stability underscores the method’s adaptability, rendering it effective for a wide range of tasks and datasets.

\begin{figure}[t]
\centering
\includegraphics[width=1\linewidth]{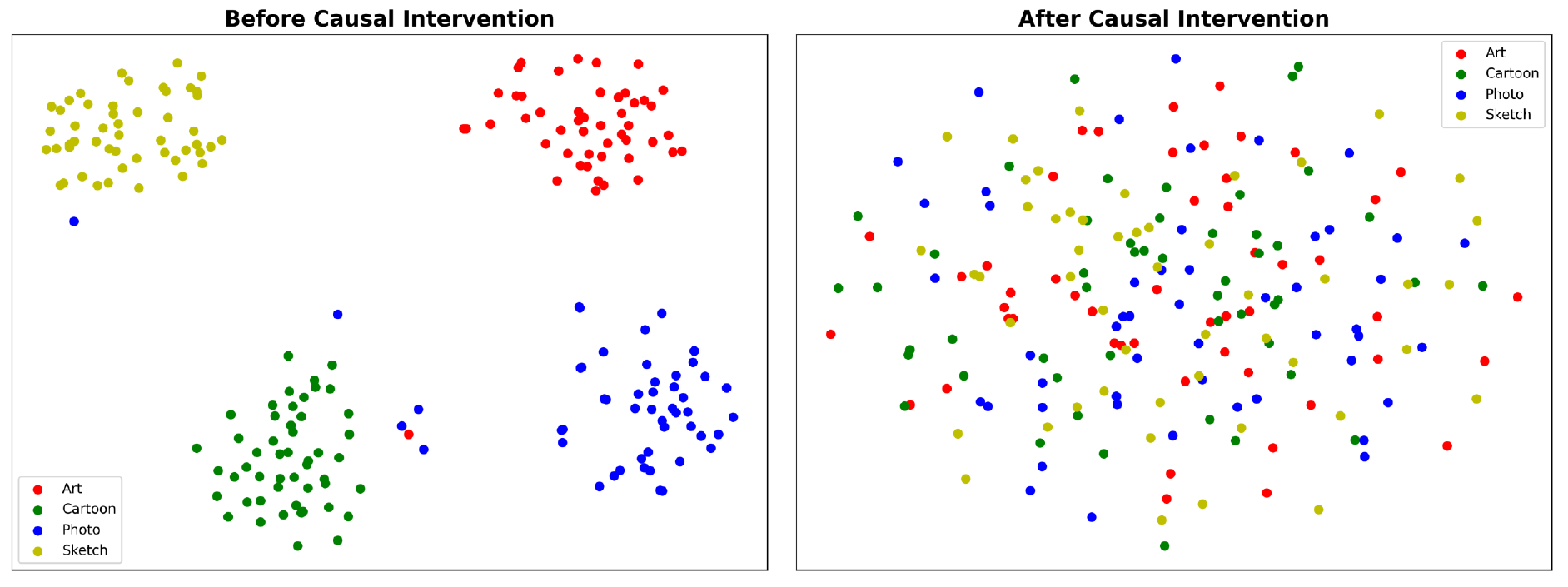}
\caption{t-SNE visualization of "dog" category features in PACS using ViT-B/16 as the visual backbone.}
\label{fig:tsne}
\end{figure}

\noindent{\bf Feature visualization.}
To further validate the effectiveness of causal intervention in learning domain-invariant features, we visualize the features of the "dog" category from the PACS dataset using t-SNE \cite{van2008visualizing} with the ViT-B/16 backbone during testing. Specifically, we compare the image features \(\bm{f}^{img}\) before intervention with the intervened features \(\mathbb{E}_{\bm{z}} [g(\bm{f}^{img}, \bm{z})]\). The visualization is conducted on test images across all four domains (Art, Cartoon, Photo, Sketch).
Results are shown in Fig.~\ref{fig:tsne}. Without intervention, the raw features \(\bm{f}^{img}\) cluster distinctly by domain, reflecting domain-specific styles (\(\bm{f}_{dom}\)) that hinder generalization. In contrast, the intervened features \(\mathbb{E}_{\bm{z}} [g(\bm{f}^{img}, \bm{z})]\) form a single cohesive cluster, with domain boundaries significantly blurred. This indicates that the intervention network \(g(\cdot)\) effectively mitigates domain-specific confounders, aligning features around class-relevant information (\(\bm{f}^{cls}\)). The shift from domain-clustered to class-clustered representations confirms that our causal intervention successfully learns domain-invariant features.

\begin{table}[h]
\centering
\caption{Ablation study on the number of layers in \(g(\cdot)\) using ViT-B/16. Accuracy (\%) and inference speed (FPS, batch size 32) are reported for each dataset and averaged across all datasets.}
\label{tab:ablation_layers}
\begin{tabular}{l|ccccc}
\toprule
Layer & 1 & 2 & 3 & 4 & 5 \\
\midrule
PACS        &95.5  &97.3  &97.7  &98.0  &98.1  \\
VLCS        &82.2  &83.9  &84.6  &84.9  &85.0  \\
Office-Home &80.8  &82.6  &83.9  &84.0  &84.3  \\
DomainNet   &58.0  &59.7  &60.4  &60.7  &61.1  \\
\midrule
Avg. &79.1  &80.9  & 81.7 &81.9  & 82.1 \\
Test Speed &13.60 &13.55 &13.46 &13.40 &13.35 \\
\bottomrule
\end{tabular}
\end{table}

\noindent{\bf Ablation Study on Layer of $g(\cdot)$.} 
We conduct an ablation study to investigate the impact of the number of layers in the causal intervention network \(g(\cdot)\), using ViT-B/16 with pre-trained CLIP weights as the backbone. The experiments are performed across four domain generalization benchmarks. The network \(g(\cdot)\) takes as input the concatenated text embedding \(\bm{f}_{\bm{s}_m}^{{cls}_k}\) and style intervention vector \(\bm{z}_n\), both of dimension \(ES\), resulting in an initial input size of \(2 \times ES\). The first layer maps this \(2 \times ES\) input to \(ES\) using a linear network, followed by ReLU activation and BatchNorm. Subsequent layers, if present, are a linear network from \(ES\) to \(ES\), each followed by ReLU and BatchNorm. We test configurations with one to five layers to determine the optimal depth for effective style intervention and generalization.
Results in Table~\ref{tab:ablation_layers} show accuracy (\%) for each dataset and the average. The one-layer baseline achieves 79.1\%, improving to 80.9\% with two layers. The three-layer setup reaches 81.7\%, balancing capacity and complexity. Beyond three layers, gains diminish (e.g., 82.1\% at five layers), suggesting limited benefit, while inference speed drops from 13.60 to 13.35 FPS. Thus, we adopt 3 layers for \(g(\cdot)\), balancing accuracy, stability, and efficiency.

\begin{table}[htbp]
\centering
\caption{Ablation study on the number of confounders \(N\) in \(Z\) using ViT-B/16.  Accuracy (\%) and inference speed (FPS, batch size 32) are reported for each dataset and averaged across all datasets.}
\label{tab:ablation_n}
\begin{tabular}{l|ccccc}
\toprule
\(N\) & 2 & 4 & 6 & 8 & 10 \\
\midrule
PACS        & 96.2 & 97.3 & 97.7 & 97.7 & 98.1 \\
VLCS        & 83.6 & 84.3 & 84.6 & 84.7 & 84.9 \\
Office-Home & 83.0 & 83.5 & 83.9 & 83.9 & 84.1 \\
DomainNet   & 59.2 & 59.7 & 60.4 & 60.7 & 60.8 \\
\midrule
Avg.        & 80.5 & 81.2 & 81.7 & 81.8 & 82.0 \\
Test Speed &13.55 &13.49 &13.46 &13.42 &13.38 \\
\bottomrule
\end{tabular}
\end{table}

\noindent{\bf Ablation Study on $N$.} 
We perform an ablation study to assess the effect of the confounder dictionary size \(N\) in TDCRL, using ViT-B/16 with pre-trained CLIP weights as the backbone across PACS, VLCS, Office-Home, and DomainNet. The confounder dictionary \(Z = \{\bm{z}_n\}_{n=1}^N\), introduced in Section \ref{sec:3.2}, comprises style intervention vectors \(\bm{z}_n\) derived from domain words, enabling causal interventions in \(g(\cdot)\). We evaluate \(N = 2, 4, 6, 8, 10\) to determine how the number of confounders impacts generalization performance.
Results are presented in Table~\ref{tab:ablation_n}, showing accuracy (\%) for each dataset and the average across all datasets. With \(N=2\), TDCRL achieves an average accuracy of 80.5\%, limited by minimal style diversity. Increasing to \(N=4\) raises this to 81.2\%, reflecting improved style coverage. The configuration with \(N=6\) reaches 81.7\%, aligning with our main results in Table \ref{tab:comparison}, and balances diversity and efficiency. Further increases to \(N=8\) and \(N=10\) yield 81.8\% and 82.0\%, respectively, but the marginal gains suggest diminishing returns due to style redundancy, while inference speed drops from 13.55 to 13.38 FPS. Thus, we select \(N=6\) for our final model, optimizing generalization without excessive complexity.

\begin{table}[htbp]
\centering
\caption{Training and inference time comparison and performance gain using ViT-B/16, measured on an RTX 4070 Ti SUPER GPU. Inference speed is reported in frames per second (FPS) with a batch size of 32.}
\label{tab:training_time}
\begin{tabular}{l|ccc}
\toprule
Method & Train Time & Test Speed& Gain (\%) \\
\midrule
CLIP & 0min & 13.92 FPS& 0.0 \\
PromptStyler & 511min & 13.79 FPS& +0.3 \\
DPStyler & 384min & 13.41 FPS& +1.2 \\
TDCRL& 255min & 13.46 FPS& +2.2 \\
\bottomrule
\end{tabular}
\end{table}

\noindent{\bf Training Time and Inference Speed.} 
We compare TDCRL’s training time and inference speed with PromptStyler \cite{cho2023promptstyler}, DPStyler \cite{tang2025dpstyler}, and CLIP (ZS-CLIP (PC) \cite{radford2021learning}), using ViT-B/16. PromptStyler’s two-stage process (prompt tuning then classifier training) takes 511 minutes. DPStyler and TDCRL use single-stage, training-free data augmentation; DPStyler trains a style removal network (384 minutes), while TDCRL trains a causal intervention network (255 minutes). CLIP requires no training.
Table~\ref{tab:training_time} also reports speed and gains on an RTX 4070 Ti SUPER GPU. Inference (batch size 32) uses CLIP’s visual encoder: PromptStyler adds a classifier (13.79 FPS), DPStyler integrates multiple models (13.41 FPS), TDCRL applies confounder and intervention (13.46 FPS), and CLIP is direct (13.92 FPS). TDCRL’s efficiency and +2.2\% gain over CLIP outperform others.


\section{Conclusion}
\label{sec:conclusion}

In this paper, we present TDCRL (Text-Driven Causal Representation Learning) for source-free domain generalization. Unlike previous methods that rely heavily on source domain data or static prompt tuning, TDCRL leverages causal inference to isolate domain-invariant features, a critical mechanism for addressing domain shifts without source access. By integrating style-augmented text embeddings with a causal intervention network, TDCRL enhances robustness and accuracy in source-free settings, particularly when test domains diverge significantly from pre-trained distributions. Our experiments on PACS, VLCS, OfficeHome, and DomainNet demonstrate that TDCRL outperforms existing CLIP-based SFDG and source-dependent methods across ResNet-50, ViT-B/16, and ViT-L/14 backbones. This makes TDCRL a practical and effective solution for addressing real-world domain generalization challenges.

\section*{Acknowledgments}
{This work was supported in part by Chinese National Natural Science Foundation Projects U23B2054, 62276254 and 62306313, the Beijing Science and Technology Plan Project Z231100005923033, Beijing Natural Science Foundation L221013, and the InnoHK program.}


\bibliographystyle{IEEEtran}
\bibliography{cite}

\vfill

\end{document}